\documentclass[sigconf]{acmart}

\copyrightyear{2026}
\acmYear{2026}
\setcopyright{cc}
\setcctype{by-nc-nd}
\acmConference[CIKM '26]{Proceedings of the 35th ACM International Conference on Information and Knowledge Management}{November 07--11, 2026}{Rome, Italy}
\acmBooktitle{Proceedings of the 35th ACM International Conference on Information and Knowledge Management (CIKM '26), November 07--11, 2026, Rome, Italy}
\acmDOI{10.1145/3799682.3840062}
\acmISBN{979-8-4007-2539-5/2026/11}

\begin{document}

\title[When Persona Simulations Are Informative]{When Persona Simulations Are Informative: Graph-Structured Signals for Pluralistic Opinion Sensing}

\author{Taehyeon An}
\affiliation{%
  \institution{Korea Advanced Institute of Science and Technology}
  \city{Seoul}
  \country{Republic of Korea}
}
\email{taeohy@kaist.ac.kr}

\author{Jaehyeong Park}
\affiliation{%
  \institution{Korea Advanced Institute of Science and Technology}
  \city{Seoul}
  \country{Republic of Korea}
}
\email{hyeong@kaist.ac.kr}

\author{Donghyuk Shin}
\authornote{Corresponding author.}
\affiliation{%
  \institution{Korea Advanced Institute of Science and Technology}
  \city{Seoul}
  \country{Republic of Korea}
}
\email{dhs@kaist.ac.kr}

\renewcommand{\shortauthors}{Taehyeon An, Jaehyeong Park, and Donghyuk Shin}

\begin{abstract}
Persona-conditioned large language models (LLMs) are increasingly used to simulate survey responses across diverse domains. However, apparent response variation can reflect unconditioned model priors or token sampling noise rather than systematic persona conditioning. We argue that persona-conditioned variation is informative when semantically similar personas exhibit concordant response shifts. To operationalize this principle, we introduce Persona-Conditioned Informativeness (PCI), an unsupervised diagnostic metric that measures whether semantically similar personas deviate in concordant directions relative to item-level sample baselines. By modeling personas as a similarity graph, PCI uses Local Moran's I to quantify local spatial coherence and extract compact persona subsets without using construct labels. To evaluate PCI without external human benchmarks, we test its ability to recover established latent value structure using the 57-item Portrait Values Questionnaire-Revised (PVQ-RR). Confirmatory factor analysis (CFA) shows that a PCI-selected 10\% subset substantially improves overall construct recovery relative to response-stability and random selection. These findings support PCI as a principled internal diagnostic for screening synthetic respondents in survey pipelines.
\end{abstract}



\maketitle

\section{Introduction}

Persona-conditioned large language models (LLMs) are increasingly used to simulate survey responses across social, commercial, and policy domains~\cite{argyleOutOneMany2023a,aroraAIHumanHybrids2025}. The goal is to study how different synthetic groups respond to diverse survey questions. Yet plausible responses are not necessarily informative: LLMs can readily generate differentiated answers even when persona information has little systematic influence on the output. For synthetic survey analysis, the central question is therefore not whether responses vary, but whether that variation is meaningfully grounded in the personas being simulated.

This challenge arises because persona profiles are typically rich and multidimensional, spanning demographics, socioeconomic status, personal values, and other attributes, whereas individual survey questions engage only particular aspects of those profiles~\cite{huQuantifyingPersonaEffect2024}. Environmental values, for example, may meaningfully shape a stance on carbon taxation, while unrelated background traits provide little consistent guidance. If an LLM effectively conditions on relevant profile cues, we should observe a coherent pattern: semantically proximate personas should shift in concordant directions rather than displaying isolated or arbitrary variation.

Existing evaluation paradigms provide limited support for identifying such coherence. Most studies evaluate synthetic samples either by comparing aggregate response distributions against external human benchmarks~\cite{argyleOutOneMany2023a} or by measuring response consistency across repeated runs~\cite{reusensAreEconomistsAlways2025}. However, macro-level alignment can mask unconditioned individual responses~\cite{bisbeeSyntheticReplacementsHuman2024}, while persona effects may differ substantially across survey items~\cite{tadaymorochoAssessingReliabilityPersonaconditioned2026}. Moreover, external human data are often unavailable during exploratory simulations or survey pre-testing. An effective internal diagnostic should therefore determine whether response variation is systematically organized with respect to persona profiles, allowing weakly grounded simulations to be identified before downstream analysis.

\begin{figure*}[t]
  \centering
  \includegraphics[width=\textwidth]{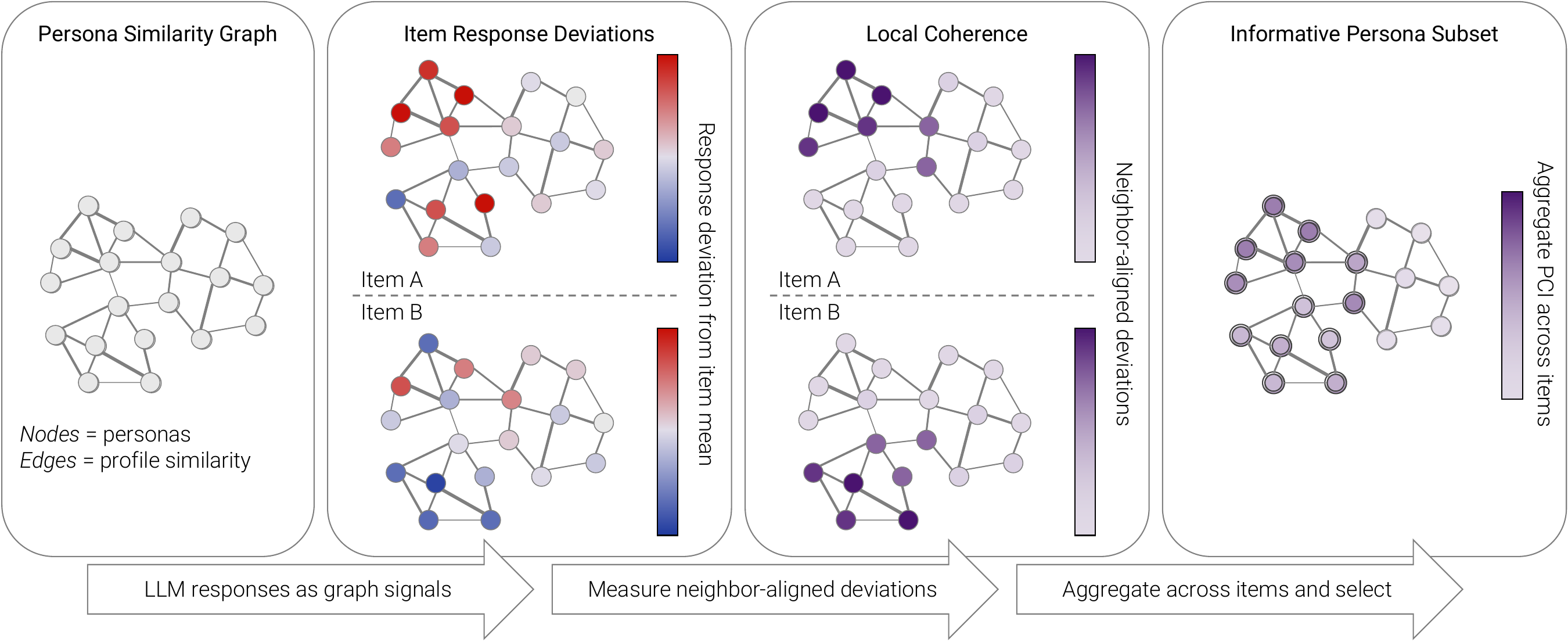}
  \caption{Overview of the PCI workflow: constructing a persona similarity graph from profile embeddings, computing centered item response deviations, scoring local spatial coherence with Local Moran's $I$, and selecting an informative persona subset.}
  \Description{Four-panel diagram of the PCI workflow: a persona similarity graph, item response deviations on the graph, local coherence scores, and the selected informative persona subset.}
  \label{fig:overview}
\end{figure*}

\begin{table*}[t]
  \caption{CFA comparison and graph sensitivity for 10\% persona subsets.}
  \label{tab:cfa}
  \centering
  \small
  \begin{tabular*}{\textwidth}{@{\extracolsep{\fill}}llrrrrr@{}}
    \toprule
    Category & Selection / graph & $\chi^2 \downarrow$ & df & CFI $\uparrow$ & SRMR $\downarrow$ & RMSEA (90\% CI) $\downarrow$ \\
    \midrule
    Baseline & Random mean (500 subsets) & 2623.78 & 1367 & .880 & .106 & .079 (.074, .084) \\
    & Response stability & 2711.10 & 1367 & .877 & .111 & .082 (.077, .086) \\
    \midrule
    Proposed & PCI, $k=740$ (50\%-NN) & 2485.15 & 1367 & \textbf{.925} & \textbf{.086} & .075 (.070, .079) \\
    \midrule
    Sensitivity & PCI, $k=444$ (30\%-NN) & \textbf{2481.24} & 1367 & .924 & .091 & \textbf{.074} (.070, .079) \\
    & PCI, $k=592$ (40\%-NN) & 2489.01 & 1367 & \textbf{.925} & .091 & .075 (.070, .079) \\
    & PCI, $k=888$ (60\%-NN) & 2633.03 & 1367 & .915 & .090 & .079 (.075, .084) \\
    & PCI, $k=1036$ (70\%-NN) & 2671.06 & 1367 & .910 & .093 & .081 (.076, .085) \\
    \bottomrule
  \end{tabular*}
  \vspace{2pt}
  \parbox{\textwidth}{\footnotesize \emph{Note.} All rows evaluate 10\% subsets ($n=148$). Random entries are means across 500 subsets. Boldface marks the best value for each metric (including ties).}
\end{table*}

To address this need, we introduce Persona-Conditioned Informativeness (PCI), an unsupervised internal diagnostic based on local spatial coherence in response deviations. Motivated by the manifold smoothness assumption~\cite{belkinManifoldRegularizationGeometric2006}, PCI models personas as a similarity graph and survey responses as graph signals. For each item, it measures whether semantically similar personas deviate from the sample-wide response baseline in concordant directions using Local Moran's $I$~\cite{anselinLocalIndicatorsSpatial1995a}, and aggregates this evidence to rank personas by informativeness. We evaluate PCI on 1,480 synthetic personas answering the 57-item Portrait Values Questionnaire-Revised (PVQ-RR)~\cite{schwartzMeasuringRefinedTheory2022a,schwartzRefiningTheoryBasic2012a}. Without using construct labels during selection, a PCI-selected 10\% subset substantially improves latent construct recovery over response-stability and random baselines ($\Delta\mathrm{CFI} = +.045$, $p=.002$), supporting local spatial coherence as a useful internal signal for screening synthetic respondents.

\section{Persona-Conditioned Informativeness}

\subsection{From Item Relevance to Local Coherence}

Figure~\ref{fig:overview} illustrates the PCI workflow. PCI operationalizes persona-conditioned informativeness through two observable properties: (1)~there must be between-persona variation around the item-level baseline, separating persona-specific variation from invariant consensus; and (2)~semantically proximate personas in profile space must deviate in concordant directions, distinguishing shared conditioning from isolated sampling variance. PCI rewards a response only when both conditions are satisfied, downweighting invariant consensus and isolated deviations that lack neighborhood support.

Formally, let $P=\{1,\dots,n\}$ denote the set of personas and $I=\{1,\dots,m\}$ the set of survey items. For persona $p$, item $i$, and generation trial $r \in \{1,\dots,R\}$, let $y_{pir}$ be the generated score. We compute the Monte Carlo mean $\bar{y}_{pi}=\frac{1}{R}\sum_{r=1}^{R}y_{pir}$ across $R$ trials to reduce stochastic token variance. While repeated-response variance is not an input to PCI, we use it to construct our response-stability baseline.

\subsection{Persona Graph and Response Signals}

Each persona possesses a narrative profile covering demographics, socioeconomic status, family structure, occupation, cultural identity, and preferences. Rather than discretizing these attributes into rigid categorical features, we embed the complete profile text with a dense representation model to capture complex, cross-attribute cues. We construct a weighted similarity graph $G=(P,E,W)$, where nodes represent personas and edges encode $k$-nearest-neighbor relations under cosine similarity. Raw edge weights are cosine similarities clipped at zero. We define the graph topology via mutual $k$-nearest-neighbor linkage and row-standardize the resulting weight matrix so that neighbor weights sum to one:
\begin{equation}
\sum_{q\in P} w_{pq}=1, \qquad w_{pq}=0 \text{ if } (p,q)\notin E.
\end{equation}
We use $k=740$ (50\%-NN for $n=1,480$) as the reference specification and evaluate sensitivity across $k \in \{30\%, \dots, 70\%\}$ in Section~\ref{sec:evaluation}. Graph density $k$ governs the extent of neighborhood support, while the selection budget controls sample selectivity.

For each item $i$, we mean-center the aggregated responses across the simulated population:
\begin{equation}
\mu_i=\frac{1}{n}\sum_{p\in P}\bar{y}_{pi}, \qquad z_{pi}=\bar{y}_{pi}-\mu_i.
\end{equation}
The item mean $\mu_i$ removes the shared sample-wide response level, making the centered deviation $z_{pi}$ capture between-persona variation around this baseline.

\subsection{Local Association and Persona Ranking}

To quantify whether $z_{pi}$ is corroborated by semantic peers, we apply Anselin's Local Moran's $I$~\cite{anselinLocalIndicatorsSpatial1995a}. Let $s_i^2=n^{-1}\sum_{p\in P}z_{pi}^{2}$ denote the second central moment of item $i$, where the zero-mean property follows from centering. When $s_i^2>0$, the local spatial association is defined as:
\begin{equation}
I_{pi}=\frac{z_{pi}\sum_{q\in P}w_{pq}z_{qi}}{s_i^2}.
\end{equation}
The weighted sum $\tilde{z}_{pi} = \sum_{q\in P}w_{pq}z_{qi}$ denotes the local spatial lag. A positive $I_{pi}$ signifies local clustering of concordant deviations: either high--high ($z_{pi}>0, \tilde{z}_{pi}>0$) or low--low ($z_{pi}<0, \tilde{z}_{pi}<0$), whereas negative $I_{pi}$ indicates spatial discordance (high--low or low--high). If $s_i^2=0$, all personas produce an identical response, yielding $I_{pi}=0$.

Because our target construct is neighbor-aligned conditioning, PCI retains only the non-negative component $\mathrm{PCI}_{pi}=\max(I_{pi},0)$. Truncating negative values ensures that discordant or isolated deviations do not contribute to informativeness, conservatively requiring neighborhood support. Aggregating across items yields the persona score $\mathrm{PCI}_{p}=m^{-1}\sum_{i\in I}\mathrm{PCI}_{pi}$. Personas with high $\mathrm{PCI}_{p}$ exhibit robust, neighbor-aligned response patterns across items, and we select the top 10\% ($n=148$) as the primary PCI subset.

\section{Construct-Recovery Evaluation}
\label{sec:evaluation}

\subsection{Data and Construct-Recovery Target}

We draw 1,480 profiles from NVIDIA's Singapore persona dataset.\footnote{\url{https://huggingface.co/datasets/nvidia/Nemotron-Personas-Singapore}} Each persona answers all 57 Portrait Values Questionnaire-Revised (PVQ-RR) items using K-EXAONE-236B-A23B.\footnote{\url{https://huggingface.co/LGAI-EXAONE/K-EXAONE-236B-A23B}} We generate $R=10$ responses per persona--item pair on a 1--6 scale and embed profiles with llama-nemotron-embed-1b-v2.\footnote{\url{https://huggingface.co/nvidia/llama-nemotron-embed-1b-v2}}

The evaluation tests whether selected responses better preserve the established PVQ-RR construct structure without using construct information during selection. The PVQ-RR measures 19 distinct human values~\cite{schwartzMeasuringRefinedTheory2022a,schwartzRefiningTheoryBasic2012a}, organized into four motivational quadrants: self-enhancement (power, achievement), self-transcendence (benevolence, universalism), openness to change (self-direction, stimulation), and conservation (security, conformity, tradition). In Schwartz's circular continuum, adjacent values represent compatible motivational goals, while opposing values represent conflicting psychological priorities. Successfully recovering this theoretical continuum through confirmatory factor analysis provides evidence that PCI enriches for structured, persona-aligned response variation rather than arbitrary subset selection.

We fit a confirmatory factor analysis (CFA) model with 19 correlated value factors and an orthogonal common response factor to control for scale acquiescence. We report the Index of Quality $\operatorname{IoQ}_v=\left|\operatorname{Corr}(C_v,\eta_v)\right|$, where $C_v$ denotes the observed composite score for value $v$ and $\eta_v$ its corresponding latent factor ($\operatorname{IoQ}_v^2$ represents the reliable composite variance). These metrics evaluate relative construct recovery across selection strategies under an identical structural model rather than absolute model fit.

\begin{table}[t]
  \caption{Index of Quality across the 19 PVQ-RR constructs.}
  \label{tab:ioq}
  \centering
  \small
  \begin{tabular*}{\linewidth}{@{\extracolsep{\fill}}lccc@{}}
    \toprule
    Value & Random & \shortstack{Response\\stability} & PCI \\
    \midrule
    Self-direction-thought & .923 & .924 & \textbf{.962} \\
    Self-direction-action & .901 & .900 & \textbf{.936} \\
    Stimulation & .974 & .977 & \textbf{.992} \\
    Hedonism & .959 & .965 & \textbf{.977} \\
    Achievement & .958 & .939 & \textbf{.983} \\
    Power-dominance & .622 & .497 & \textbf{.758} \\
    Power-resources & .935 & .913 & \textbf{.971} \\
    Face & .793 & .711 & \textbf{.831} \\
    Security-personal & .699 & .713 & \textbf{.811} \\
    Security-societal & .875 & \textbf{.897} & \textbf{.897} \\
    Tradition & .949 & .938 & \textbf{.979} \\
    Conformity-rules & .945 & .952 & \textbf{.974} \\
    Conformity-interpersonal & .948 & .936 & \textbf{.979} \\
    Humility & .902 & .879 & \textbf{.962} \\
    Benevolence-care & .709 & .746 & \textbf{.842} \\
    Benevolence-dependability & .536 & \textbf{.649} & .480 \\
    Universalism-concern & .840 & \textbf{.893} & .791 \\
    Universalism-nature & .916 & \textbf{.928} & .911 \\
    Universalism-tolerance & .904 & .916 & \textbf{.939} \\
    \midrule
    Mean & .857 & .856 & \textbf{.893} \\
    Median & .904 & .913 & \textbf{.939} \\
    \bottomrule
  \end{tabular*}
  \vspace{2pt}
  \parbox{\linewidth}{\footnotesize \emph{Note.} All values use 10\% subsets. Random entries are means across 500 subsets. Boldface marks the best value in each row (including ties).}
\end{table}

\begin{figure}[t]
  \centering
  \includegraphics[width=\linewidth]{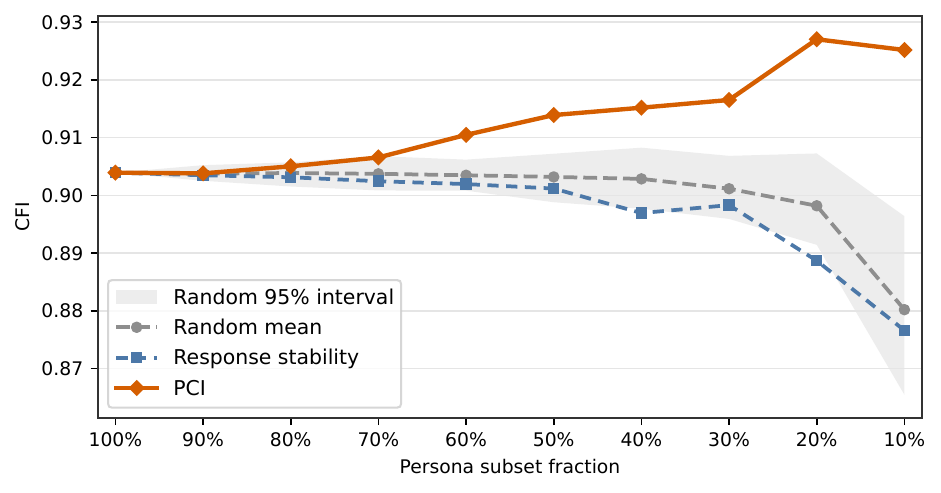}
  \caption{CFI across subset sizes with the 50\%-NN graph fixed. The 10\% random subset distribution contains 500 subsets.}
  \Description{Line chart of CFI by subset fraction for PCI, response stability, and random selection, where PCI achieves the highest CFI from 10\% through 80\% subsets.}
  \label{fig:sweep}
\end{figure}

\subsection{Selection Rules and Statistical Tests}

We compare three 10\% subsets ($n=148$): PCI selects the highest $\mathrm{PCI}_{p}$ scores; response stability selects the lowest generation variance $V_p=m^{-1}\sum_i\operatorname{Var}_r(y_{pir})$; and the random baseline uses 500 size-matched random subsets as a chance-level reference for the selection strategies. We evaluate graph density across $k\in\{30\%, \dots, 70\%\}$-NN around the reference 50\%-NN specification, alongside subset fractions swept from 10\% to 100\% to isolate budget and density effects. We compute empirical one-sided $p$-values against the 500 size-matched random subsets and apply Bonferroni correction across the five graph-density specifications.

\subsection{Results}

As shown in Table~\ref{tab:cfa}, the primary 10\% PCI subset achieves a higher CFI (.925) than the response-stability baseline (.877) and significantly exceeds the 500-subset random baseline mean of .880 (95\% empirical random-subset interval: .865--.896; $p=.002$, Bonferroni-adjusted $p=.010$). PCI similarly improves residual fit, reducing SRMR from .106 to .086 ($p=.002$, adjusted $p=.010$) alongside consistent reductions in $\chi^2$ and RMSEA.

As reported in Table~\ref{tab:ioq}, construct-level evaluations reveal consistent advantages alongside construct-specific nuances. PCI attains the best or tied-best IoQ for 16 of 19 constructs, achieving an average of $.893$ compared to $.857$ for random selection and $.856$ for response stability. Gains are particularly pronounced for power-dominance (from .622 to .758) and security-personal (from .699 to .811). Conversely, response stability performs better on several constructs, including benevolence-dependability and two universalism constructs, indicating that generation consistency and persona-conditioned local coherence capture distinct properties of synthetic responses.

Across subset fractions, PCI consistently outperforms both baselines, maintaining superior CFI from 10\% through 80\% fractions and peaking at .927 at a 20\% fraction ($n=296$, Figure~\ref{fig:sweep}). In addition, sensitivity analysis demonstrates that these improvements remain robust across graph densities: CFI forms a stable plateau of $.924$--$.925$ between 30\% and 50\%-NN, remains elevated at $.910$--$.915$ up to 70\%-NN, and preserves highly consistent persona rankings throughout (Spearman $\rho \in [.961, .990]$).

\section{Discussion and Conclusion}

In computational social science, survey pre-testing, and pluralistic opinion sensing, plausible and repeatable outputs can mask unconditioned model consensus. PCI provides an unsupervised internal diagnostic that quantifies whether item-level response deviations exhibit local alignment among semantically proximate personas, without requiring construct labels. The observed construct-recovery gains support local spatial coherence as a useful internal signal of structured response variation in synthetic respondents.

For research and practical applications, PCI suggests a three-stage screening workflow: (1)~unsupervised auditing of raw synthetic generations to flag items exhibiting weak persona-aligned structure, (2)~subset selection to extract compact cohorts that retain structurally informative variation for cost-effective exploratory surveys, and (3)~targeted validation triage directing costly human validation toward items or subgroups exhibiting low informativeness or discordance. Rather than sampling personas blindly, this workflow concentrates auditing effort where needed. Because PCI is closed-form given fixed generations and graph construction, it remains computationally lightweight without requiring model training.

Several limitations qualify our findings and highlight promising directions for future research. First, internal construct recovery provides evidence of coherent profile--response structure, but serves as an internal structural diagnostic rather than a guarantee of external population validity; combining PCI with external calibration remains essential for representative polling. Second, our primary graph construction embeds full narrative profiles into a single global metric space. Because survey items may engage different subsets of attributes, constructing subspace-specific or multi-relational persona graphs represents a promising direction to prevent irrelevant profile cues from diluting local neighborhood structures. Finally, while demonstrated on the 57-item Portrait Values Questionnaire-Revised with K-EXAONE-236B, evaluating cross-model generalizability across diverse model families and multilingual instruments will further delineate the boundaries of synthetic opinion sensing.

\begin{acks}
This work was supported by the \grantsponsor{IITP}{Institute of Information \& Communications Technology Planning \& Evaluation (IITP)}{https://www.iitp.kr}-Global Data-X Leader HRD program grant funded by the \grantsponsor{MSIT}{Korea government (MSIT)}{https://www.msit.go.kr} (\grantnum{IITP}{IITP-RS-2024-00440626}).
\end{acks}

\section*{GenAI Disclosure}
Generative AI supported translation, editing, and \LaTeX{} formatting.

\bibliographystyle{ACM-Reference-Format}
\balance
\bibliography{references}

\end{document}